\documentclass{article}

\usepackage[siunitx]{circuitikz}
\usetikzlibrary{positioning}
\usepackage{tikz}
\usepackage{amsmath}
\usepackage{graphicx}

\usepackage{spconf}
\usepackage{graphicx}
\usepackage{mathtools} % loads amsmath
\usepackage{amssymb,amsfonts}

\usepackage[ruled,vlined]{algorithm2e}

\usepackage{booktabs}
\usepackage{multirow}
\usepackage{tabularx,array}

\usepackage{siunitx}
\usepackage{bm}
\usepackage[nolist]{acronym}
\usepackage{enumitem}
\usepackage{xcolor}
\usepackage{cite}

\usepackage{hyperref}

\DeclareSIUnit{\var}{VAr}
\DeclareSIUnit{\voltampereReactive}{VAr}
\DeclareSIUnit{\ohm}{\text{\ensuremath{\Omega}}}
\newcolumntype{Y}{>{\raggedleft\arraybackslash}X}

\def\BibTeX{{\rm B\kern-.05em{\sc i\kern-.025em b}\kern-.08em
    T\kern-.1667em\lower.7ex\hbox{E}\kern-.125emX}}

\begin{document}
% --- Acronyms (define once; use \ac{<key>} in text) ---

\newacro{pignn}[PIGNN]{Physics-informed graph neural network}
\newacro{nr}[NR]{Newton--Raphson}
\newacro{pf}[PF]{Power Flow}
\newacro{mv}[MV]{Medium Voltage}
\newacro{hv}[HV]{High Voltage}

\newacro{ac}[AC]{Alternating Current}
\newacro{dc}[DC]{Direct Current}

\newacro{acpf}[AC-PF]{Alternating-Current Power Flow}

\newacro{fdlf}[FDLF]{Fast-Decoupled Load Flow}
\newacro{distflow}[DistFlow]{Backward/Forward Sweep}

\newacro{mvn}[MVN]{Medium-Voltage Network}
\newacro{hvn}[HVN]{High-Voltage Network}
\newacro{hmv}[HMV]{Joint HV+MV}

\newacro{pv}[PV]{Generator-Voltage (PV) Bus}
\newacro{pq}[PQ]{Load (PQ) Bus}

\newacro{gnn}[GNN]{Graph Neural Network}
\newacroplural{gnn}[GNNs]{Graph Neural Networks}
\newacro{pinn}[PINN]{Physics-Informed Neural Network}

\newacro{mlp}[MLP]{Multi-Layer Perceptron}
\newacro{gat}[GAT]{Graph Attention Network}
\newacro{gcn}[GCN]{Graph Convolutional Network}

\newacro{ls}[LS]{Line Search}
\newacro{rmse}[RMSE]{Root Mean Squared Error}
\newacro{bfs}[BFS]{Breadth-First Search}
\newacro{iqr}[IQR]{Interquartile Range}
\newacro{lr}[LR]{Learning Rate}

\newacro{cpu}[CPU]{Central Processing Unit}
\newacro{gpu}[GPU]{Graphics Processing Unit}
\newacro{blas}[BLAS]{Basic Linear Algebra Subprograms}

\newacro{pu}[p.u.]{per-unit}

% \title{Adaptive Physics-Guided Graph Neural Solver for MV/HV AC Load-Flow with Residual-Driven Mini-Batching}
% \title{Physics-Informed Graph Neural Network with Attention and Line Search for Medium-High Voltage AC Power Flow}
\title{Physics-Informed GNN for Medium-High Voltage AC Power Flow with Edge-Aware Attention and Line Search Correction Operator}

\name{%
\begin{tabular}{@{}c@{}}
Changhun Kim$^{1}$\thanks{Corresponding author: \texttt{changhun.kim@fau.de}. This work was conducted within the scope of the research project \textit{GridAssist} and was supported through the “OptiNetD” funding initiative by the German Federal Ministry for Economic Affairs and Energy (BMWE) as part of the 8\textsuperscript{th} Energy Research Programme.},
Timon Conrad$^{2}$, Redwanul Karim$^{1}$, Julian Oelhaf$^{1}$, David Riebesel$^{2}$, \\
Tomás Arias-Vergara$^{1}$, Andreas Maier$^{1}$, Johann Jäger$^{2}$, Siming Bayer$^{1}$
\end{tabular}
}

\address{
$^{1}$ Pattern Recognition Lab, Friedrich-Alexander-Universität Erlangen-Nürnberg, Germany \\
$^{2}$ Institute of Electrical Energy Systems, Friedrich-Alexander-Universität Erlangen-Nürnberg, Germany
}

\ninept
\maketitle

\begin{abstract}
Physics-informed graph neural networks (PIGNNs) have emerged as fast AC power-flow solvers that can replace the classic Newton--Raphson (NR) solvers, especially when thousands of scenarios must be evaluated. However, current PIGNNs still need accuracy improvements at parity speed; in particular, the soft constraint on the physics loss is inoperative at inference, which can deter operational adoption. We address this with PIGNN-Attn-LS, combining an edge-aware attention mechanism that explicitly encodes line physics via per-edge biases to form a fully differentiable known-operator layer inside the computation graph, with a backtracking line-search-based globalized correction operator that restores an operative decrease criterion at inference. Training and testing use a realistic High-/Medium-Voltage scenario generator, with NR used only to construct reference states. On held-out HV cases consisting of \(4\)--\(32\)-bus grids, PIGNN-Attn-LS achieves a test RMSE of  \(0.00033\) p.u. in voltage and \(0.08^{\circ}\) in angle, outperforming the PIGNN-MLP baseline by 99.5\% and 87.1\%, respectively. With streaming micro-batches, it delivers 2--5\(\times\) faster batched inference than NR on \(4\)--\(1024\)-bus grids.

\end{abstract}

\begin{keywords}
AC power flow, physics-informed neural networks, known operator learning, graph neural networks, backtracking line search
\end{keywords}

\section{Introduction}
AC \ac{pf} computes bus voltages and angles that are consistent with network physics. It serves as a cornerstone for critical operational and planning tasks, including security assessment and optimization tasks\cite{ElFergany2025LoadFlowReview,SKOLFIELD2022387,HAILU2023e14524}. In real-world practice, system operators must repeatedly solve AC-PF across large scenario sets spanning both \ac{hv} and \ac{mv} grids. This creates two central requirements: robust convergence across diverse operating conditions and high computational throughput for large-scale scenario evaluation.

State of the practice PF solvers only partly meet these needs. \ac{nr} often converges rapidly when well initialized, but calculating inverse of Jacobian via sparse triangular solves is not naturally GPU-friendly \cite{4073219}. Fast-decoupled load flow (FDLF) is quicker but relies on reactance dominant, weak-coupling assumptions and loses accuracy when those assumptions break \cite{stott1974fdlf}. In both approaches, performance hinges on large sparse linear systems where cost worsens with network size and conditioning\cite{tinney1967sparse}, creating a brittle accuracy–speed trade-off: improving the robustness of the solver often slows it down, whereas making it fast typically degrades accuracy.

Plain ML/DL surrogates \cite{ElFergany2025LoadFlowReview} are fast but tied to a fixed topology and therefore do not scale, while graph-based GNN surrogates \cite{deihim2024initialestimate} transfer across sizes and layouts, but are typically not physics-informed because they imitate NR outputs. In contrast, \acp{pignn} combine physics guidance with graph scalability: they model the grid as a graph and learn a residual-to-update operator that maps power-mismatch signals \(\Delta P\) and \(\Delta Q\) to voltage--angle corrections via message passing, making them a strong choice for large, variable grids.

\begin{figure}[t]
    \centering
    % Control both width and height individually
    \resizebox{0.48\textwidth}{0.38\textheight}{%
        \input{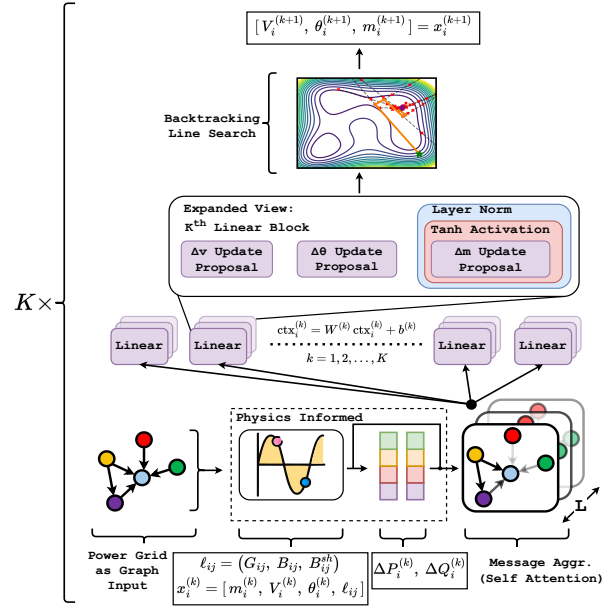}
    }
    \caption{Architecture of PIGNN-Attn-LS. Message aggregation uses L layers of edge-aware multi-head self-attention and is interchangeable with an MLP. The central node denotes the bus of interest; incoming edges are weighted by attention scores, with lower opacity indicating lower importance.}
    \label{fig:my_figure}
\end{figure}

Trained with physics losses rather than NR labels, PIGNNs amortize computation over scenarios, preserve inductive biases from Kirchhoff's laws, and can generalize across grid sizes. Properly designed, a PIGNN behaves like a learned inverse Jacobian, producing corrections that either (i) solve AC-PF directly or (ii) warm-start NR so that only a few Newton steps are needed, maintaining NR precision while reducing wall time.

Prior PIGNN research has established a strong foundation along three practical lines: (i) \emph{end-to-end physics-informed solvers}, where unrolled message passing trained on residual penalties delivers high throughput with modest memory, spanning node-centric and bidirectional node/edge updates to shallow GCNs \cite{DONON2020106547,boettcher2023acpfgnn,dejongh2022pigdl,lin2023powerflownet,yang2022mfnn,lopezgarcia2022typedgnn,Owerko2020OPF_GNN}; (ii) \emph{stronger physics or hybrid constraints}, in which physics-guided decoders and augmented-Lagrangian objectives tighten the learning signal and accommodate uncertain admittances and mixed AC/DC grids within a single pipeline \cite{hu2021physicsguidedpf,yang2023pggnn}; and (iii) \emph{warm starts for numerical optimization}, where learned initializers reduce downstream iterations; RL- and GNN-based warm starts shorten Newton steps and accelerate optimal power flow on various benchmarks \cite{deihim2024initialestimate,diehl2019warmstart,yan2025datanrinit}.

Building on this foundation, the pursuit of accuracy at parity speed remains a central research direction. A recurring gap is that \textit{physics-informed} often enters only through the training objective: as a regularizer, it penalizes violations of the physical equations during training, but it does not provide a mechanism that guarantees physically plausible behavior at inference time \cite{Maier2022KOL,Krishnapriyan2021PINNFailureModes}. Motivated by known-operator learning \cite{Maier2022KOL}, we inject admittance-derived coupling into message passing via attention biases, yielding an explicit, differentiable known-operator layer inside the computational graph rather than treating admittances as additional input features.

To further operationalize this bias during inference, we augment the unrolled solver with a backtracking line search \cite{Armijo1966MinimizationOF} on a power-mismatch merit function. It adaptively selects the step size $\alpha$ for each of the $K$ updates in $(\Delta V,\Delta\theta)$, enforcing sufficient decrease and improving robustness under out-of-distribution operating conditions.

For evaluating the proposed model, the choice of data is crucial. Although various benchmark power grid data sets \cite{lovett2024opfdata, bhagavathula2025pfdelta} and synthetic data generation methods \cite{hamann2024foundationgridfm} have recently been released, \emph{we build our own scenario synthesis pipeline} designed to better reflect European and German grids.

Finally, \emph{we analyze throughput} in single-case and multi-scenario settings, comparing CPU Newton--Raphson (multiprocessing) with GPU PIGNN using streaming micro-batches. We report median inference over warm-up and repeated runs and identify the problem-size regimes where PIGNN surpasses NR.

Code: \href{https://github.com/Kimchangheon/PIGNN-Attn-LS}{github.com/Kimchangheon/PIGNN-Attn-LS}

\section{Methodology}

\subsection{Problem Formulation}
\label{ProblemFormulation}
We model the grid as an undirected graph $\mathcal{G}=(\mathcal{V},\mathcal{E})$ with $N=|\mathcal{V}|$ buses. The node state is complex voltage
$x=[V,\,\theta]\in\mathbb{R}^{2N}$, where $V=|\underline{V}|$ and $\underline{V}=V\cdot e^{\jmath \theta}$. Currents are $\underline{I}=\underline{Y}\cdot\underline{V}$ with $\underline{Y}\in\mathbb{C}^{N\times N}$ and complex powers $\underline{S}=\underline{V}\cdot\underline{I}^*$, giving $P=\Re\{\underline{S}\}$ and $Q=\Im\{\underline{S}\}$. For setpoints $(P^{\mathrm{set}},Q^{\mathrm{set}})$, define residuals
\[
\Delta P = P^{\mathrm{set}} - P, \qquad
\Delta Q = Q^{\mathrm{set}} - Q.
\]

Classical Newton--Raphson solves PF by computing the step $\Delta x=-J(x)^{-1} r(x)$ for $r(x^\star)=0$, where
\[
r(x)=\begin{bmatrix}\Delta P\\ \Delta Q\end{bmatrix},\quad
J=
\begin{bmatrix}
\frac{\partial P}{\partial \theta} & \frac{\partial P}{\partial V}\\
\frac{\partial Q}{\partial \theta} & \frac{\partial Q}{\partial V}
\end{bmatrix}
=
\begin{bmatrix}
H & N\\
M & L
\end{bmatrix}.
\]

\({J}\) depends nonlinearly on admittances, voltages, and angle differences; building and solving it is expensive on large meshed grids\cite{faucris}.

Instead, we learn a graph-structured residual-to-update operator
\[
\Delta x \approx\mathcal{G}_\theta\!\big(x,Y)\,r(x).
\]

\subsection{Unrolling \(K\) Correction Steps}
We unroll \(K\) iterative correction steps of message passing over \(\mathcal{G}\), producing a sequence \(\{x^{(k)}\}_{k=0}^{K}\). At each step \(k\), we compute mismatch residuals \(\big(\Delta P^{(k)}, \Delta Q^{(k)}\big)\), exchange messages among neighboring nodes, and propose updates \(\big(\Delta \theta^{(k)}, \Delta V^{(k)}\big)\). 
The generic update at node \(i\) is : 
\begin{multline}
m_i^{(k+1)} \;=\; \operatorname{UPDATE}^{(k)}\!\Big(
m_i^{(k)},\;
\Phi_{\text{phys}}^{(k)} \!\Big( \\
\operatorname{AGGREGATE}^{(k)}\!\left(\{\,m_j^{(k)} \mid j \in \mathcal{N}(i)\}\right),
\text{state}^{(k)}\Big)\Big).
\end{multline}
where \(\Phi_{\text{phys}}^{(k)}\) injects the residuals into the node features, forming a physically grounded learning signal that guides the neural solver toward satisfying the AC--PF equations in ~\ref{ProblemFormulation}. Accordingly, training minimizes a discounted sum of stepwise-mismatch penalties over the unrolled iterations, with discount factor \(\gamma\in(0,1]\):
\[
\mathcal{L}_{\mathrm{phys}}
= \sum_{k=0}^{K-1} \gamma^{K-1-k}\,\frac{1}{N}\sum_{i=1}^{N}
\big[(\Delta P_i^{(k)})^2+(\Delta Q_i^{(k)})^2\big].
\]

Building on this idea, we construct a baseline, PIGNN-MLP, which follows the Graph Neural Solver paradigm \cite{DONON2020106547} and uses an multi-layer perceptron (MLP) message aggregator in the DeepSets sense \cite{zaheer2017deepsets}. This choice makes neighbor aggregation permutation invariant and, despite its simplicity, yields competitive performance. Nevertheless, the resulting operator behaves as a stationary, isotropic graph filter. However, AC power-flow sensitivities are state dependent, directional, and strongly influenced by the line admittances \(Y_{ij}\). Uniform aggregation therefore cannot emulate the anisotropic structure of \(J(x)^{-1}\), motivating a more expressive, physics-aware formulation.

\subsection{Physics-Informed, Edge-Aware Multi-Head Self-Attention}
\label{sec:attn-agg}
To overcome the limitations of isotropic MLP aggregation, we adopt a Transformer-style \cite{NIPS2017_3f5ee243}, edge-aware multi-head self-attention aggregator. The physics-informed node features by \(\Phi_{\text{phys}}^{(k)}\) at iteration \(k\),
\(x_i^{(k)}=\left[V_i^{(k)}, \theta_i^{(k)}, \Delta P_i^{(k)}, \Delta Q_i^{(k)}, m_i^{(k)}\right] \in \mathbb{R}^{4+d}\), we then form queries, keys, and values at step \(k\). For each undirected edge \((i,j)\in\mathcal{E}\), we consider both scored directions \(j \to i\) and \(i \to j\). For head \(h=1,\ldots,H\) with per-head dimensionality \(d_h=d_{\text{model}}/H\),
\[
q_i^{(h)} = W_Q^{(h)} x_i^{(k)}, \qquad
k_j^{(h)} = W_K^{(h)} x_j^{(k)}, \qquad
v_j^{(h)} = W_V^{(h)} x_j^{(k)}.
\]

The attention score is computed as a softmax over the neighbors
\[
\alpha_{ij}^{(h)} = \operatorname{softmax}\!\left( \frac{\left\langle q_i^{(h)}, k_j^{(h)} \right\rangle}{\sqrt{d_h}} + \beta_{ij}^{(h)} \right),
\qquad 
\beta_{ij}^{(h)} = f_{\mathrm{edge}}^{(h)}\!\left(\ell_{ij}\right)
\]
and the multi-head aggregated context delivered to node $i$ is:
\[
\operatorname{ctx}_i^{(k)}
= W_O\!\left[
\sum_{j}\alpha_{ij}^{(1)}v_j^{(1)}\;\big\|\;\cdots\;\big\|\;\sum_{j}\alpha_{ij}^{(H)}v_j^{(H)}
\right]\in\mathbb{R}^{d}.
\]

% softmax over the incoming edges of each destination node, done separately for every batch b and head h—all in one vectorized pass.

% The edge-dependent bias \(\beta_{ij}^{(h)}\) injects admittance features directly into the attention scores, yielding a strong line-physics inductive bias; by contrast, MLP or standard graph attention network \cite{GAT} aggregation incorporates admittance only implicitly via node features \(\Delta P,\Delta Q\).

% The edge-dependent bias \(\beta_{ij}^{(h)}\) injects admittance features directly into the attention scores, yielding a strong line-physics inductive bias. From a known-operator learning viewpoint \cite{Maier2022KOL}, \(\beta_{ij}^{(h)}\) acts as a lightweight, fully differentiable \emph{known-operator layer} inside the computation graph: it encodes the physically known edge coupling in the forward pass rather than leaving it to be learned implicitly. By contrast, MLP or standard graph attention network \cite{GAT} aggregation incorporates admittance only implicitly via node features \(\Delta P,\Delta Q\).

The edge-dependent bias \(\beta_{ij}^{(h)}\) injects admittance features into attention scores. From a known-operator learning viewpoint \cite{Maier2022KOL}, it acts as a lightweight, fully differentiable known-operator layer that makes edge coupling explicit in the forward pass, unlike MLP/GAT-style aggregation \cite{GAT}, where it is only implicit via node features.

Leveraging this explicit encoding of line physics, the mechanism (i) yields a state-dependent, non-stationary propagation operator \(\Pi_h[i,j]=\alpha_{ij}^{(h)}(x,Y)\) that can approximate the local inverse-Jacobian action \(\Delta x \approx \mathcal{G}_\theta(x,Y)\,r\) more faithfully than uniform aggregators; (ii) by duplicating each undirected edge into two scored directions, it allows \(\alpha_{ij}\neq\alpha_{ji}\), matching the directionality in the Jacobian blocks (\(H\neq H^\top\) in general); (iii) the multi-head structure lets different heads specialize to, for example, conductive vs.\ susceptive couplings, forming a low-rank, block-structured approximation to \(J(x)^{-1}\); and (iv) stacking \(L\) attention layers per iteration yields within-iteration multi-hop context without increasing \(K\). Conceptually and empirically, this physics-aware attention improves power-flow learning relative to uniform aggregation.

\subsection{Backtracking line search and Voltage update caps}
In nonlinear optimization, globalization techniques such as backtracking line search\cite{Armijo1966MinimizationOF} enforce descent by adaptively selecting the step size to guarantee a sufficient decrease of a merit function. To compensate for the inoperative physics loss at inference, we apply a backtracking line search at each message-passing iteration $k$ using the power-mismatch merit function. This restores an operative decrease criterion during inference and improves stability. The detailed procedure appears in Algorithm~\ref{alg:armijo-pignn}. In addition, we enforce per-iteration step caps and voltage bounds to promote monotonic decrease and to keep states within realistic operating ranges.

\begin{algorithm}[htbp]
  \caption{Backtracking Line Search with Update Caps}
  \label{alg:armijo-pignn}
  \SetKwInOut{Input}{Input}\SetKwInOut{Output}{Output}
  \Input{Current state \((\theta^{(k)}, V^{(k)}, m^{(k)})\); \\
  proposed updates \((\Delta\theta^{(k)}, \Delta V^{(k)}, \Delta m^{(k)})\);\\ constants \(c_1\), \(\rho\), \(\alpha_{\min}\), \(d_\theta^{\max}\), \(d_v^{\mathrm{frac}}\), \(V_{\min}\), \(V_{\max}\).}
  \Output{Updated state \((\theta^{(k+1)}, V^{(k+1)}, m^{(k+1)})\).}
  \textbf{Merit function:} \(F(V,\theta)=\max\!\left\{\|\Delta P(V,\theta)\|_{\infty},\ \|\Delta Q(V,\theta)\|_{\infty}\right\}\).\;

  % Step caps:
  Clip \(\Delta\theta^{(k)}\) and \(\Delta V^{(k)}\) to per-step caps \\ \(|\Delta\theta^{(k)}|\le d_\theta^{\max}\), \(|\Delta V^{(k)}|\le d_v^{\mathrm{frac}}\, V^{(k)}\).\;
  Ensure state within bounds: \\ \(\theta^{(k)}\leftarrow \mathrm{wrap}(\theta^{(k)})\), \(V^{(k)}\leftarrow \mathrm{clip}(V^{(k)})\).\;

  Initialize \(\alpha \leftarrow 1\); set \(F_k \leftarrow F(V^{(k)}, \theta^{(k)})\).\;

  \While{\(F\!\bigl(\mathrm{clip}(V^{(k)}+\alpha\,\Delta V^{(k)}),\, \mathrm{wrap}(\theta^{(k)}+\alpha\,\Delta\theta^{(k)})\bigr) > (1-c_1 \alpha)\,F_k\)}{
    \(\alpha \leftarrow \rho\,\alpha\) \tcp*{Backtrack}
    \If{\(\alpha < \alpha_{\min}\)}{\textbf{break}}
  }

  \eIf{\(\alpha \ge \alpha_{\min}\)\ \textbf{or}\ \(F\!\bigl(\mathrm{clip}(V^{(k)}+\alpha_{\min}\Delta V^{(k)}),\, \mathrm{wrap}(\theta^{(k)}+\alpha_{\min}\Delta\theta^{(k)})\bigr) < F_k\)}{
    {\small \(\theta^{(k+1)} = \mathrm{wrap}\!\bigl(\theta^{(k)} + \alpha\, \Delta\theta^{(k)}\bigr), 
               V^{(k+1)} = \mathrm{clip}\!\bigl(V^{(k)} + \alpha\, \Delta V^{(k)}\bigr),  
               m^{(k+1)} = m^{(k)} + \alpha\, \Delta m^{(k)}\)}\;
  }{
    {\small \(\theta^{(k+1)} = \theta^{(k)},\quad
               V^{(k+1)} = V^{(k)},\quad
               m^{(k+1)} = m^{(k)}\)}\;
  }
\end{algorithm}

\section{Scenario Synthesis and Reference Power-Flow Solutions (MV \& HV)}

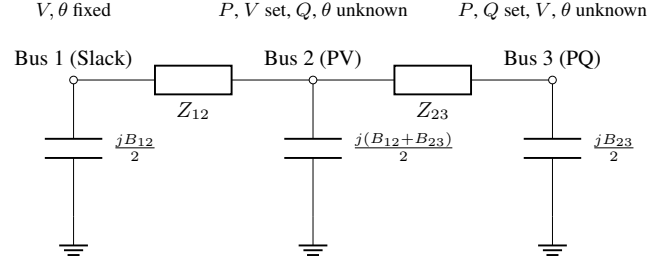
\begin{figure}[t]
\centering
\resizebox{\linewidth}{!}{%
\begin{circuitikz}
  %--- bus nodes --------------------------------------------------------------
  \draw
    (0,0) node[ocirc,label=above:{\small Bus 1 (Slack)}] (B1) {}
    (3.5,0) node[ocirc,label=above:{\small Bus 2 (PV)}]    (B2) {}
    (7,0) node[ocirc,label=above:{\small Bus 3 (PQ)}]    (B3) {};
  %--- series branches (use a generic box to mean Z = R + jX) -----------------
  \draw (B1) to[generic,l_={$Z_{12}$}] (B2)
        (B2) to[generic,l_={$Z_{23}$}] (B3);
  %--- shunt charging (π-model) -----------------------------------------------
  \draw (B1) to[C,l={$\,\tfrac{jB_{12}}{2}$}] (0,-2) node[ground]{};
  \draw (B2) to[C,l={$\,\tfrac{j(B_{12}+B_{23})}{2}$}] (3.5,-2) node[ground]{};
  \draw (B3) to[C,l={$\,\tfrac{jB_{23}}{2}$}] (7,-2) node[ground]{};
  %--- small callouts for bus constraints -------------------------------------
  \node at (0,1.0)  {\footnotesize $V,\theta$ fixed};
  \node at (3.5,1.0)  {\footnotesize  $P$, $V$ set,  $Q$, $\theta$ unknown};
  \node at (7,1.0)  {\footnotesize $P$, $Q$ set, $V$, $\theta$ unknown};
\end{circuitikz}%
}
\caption[Three-bus $\pi$-model]{\small Three-bus $\pi$-model (series $R{+}jX$; half shunt per line end). 
$Z_{ij}=(R'+jX')L_{ij}$, $y_{ij}=1/Z_{ij}$, $B^{\mathrm{sh}}_{ij}=\omega C' L_{ij}$ (total per line, split as $B^{\mathrm{sh}}_{ij}/2$ at each end); 
$B^{\mathrm{sh}}_{i}=\tfrac{1}{2}\!\sum_{j\in\mathcal{N}(i)}\!B^{\mathrm{sh}}_{ij}$. 
Admittance matrix: $Y_{ij}=-y_{ij}$ on lines and $Y_{ii}=\sum_{k\in\mathcal{N}(i)} y_{ik}+\jmath B^{\mathrm{sh}}_{i}$.}

\label{fig:threebus}
\end{figure}

Guided by Europe/Germany–typical overhead-line parameter ranges (Table~\ref{tab:distribution}) for series impedance and shunt charging \cite{oswald2005eev,kremens1996,theil2012}, we synthesize physically plausible MV/HV grids, instantiating each line with a $\pi$-model; the series and shunt terms are indicated in Fig.~\ref{fig:threebus}. For each case, we draw a connected topology with one Slack bus and the remaining buses typed as PV or PQ, and we sample operating injections $(P_i^{\mathrm{set}},Q_i^{\mathrm{set}})$ by regime. Initial voltages $V^{(0)}$ use magnitudes in $[0.9,1.1]$~$\mathrm{p.u.}$ for Slack and PV buses and $1.0$~$\mathrm{p.u.}$ for PQ buses, with zero angle; conversion to engineering units follows $V_{\text{base}}$. We then compute a reference solution by running NR for up to $K$ iterations on $(Y,S^{\mathrm{set}})$ under Slack/PV/PQ constraints to obtain $(V^\ast,S^\ast)$. Non-convergent or disconnected draws are discarded. MV cases emphasize higher $R/X$ and shorter spans; HV cases are reactance-dominated with longer spans.
\vspace{-0.5\baselineskip}

\begin{table}[h]
\centering
\caption{Parameter ranges by MV/HV regime (engineering units).}
\label{tab:distribution}
\begin{tabular}{|l|c|c|}
\hline
\textbf{Parameter} & \textbf{MV} & \textbf{HV} \\
\hline
Grid voltage $V_{base}$               & 10\,kV        & 110\,kV \\
Base power $S_{base}$                 & 10\,MVA       & 100\,MVA \\
Line length $L$                     & 1--20\,km     & 1--50\,km \\
Series resistance $R'$          & 0.5--0.6\,$\Omega$/km & 0.15--0.2\,$\Omega$/km \\
Series reactance $X'$           & 0.3--0.35\,$\Omega$/km & 0.35--0.45\,$\Omega$/km \\
Shunt capacitance $C'$          & 8--14\,nF/km  & 8--10\,nF/km \\
Active power $P$      & $[-5,\,5]\ \mathrm{MW}$            & $[-300,\,300]\ \mathrm{MW}$ \\
Reactive power $Q$    & $[-2,\,2]\ \mathrm{MVAr}$          & $[-150,\,150]\ \mathrm{MVAr}$ \\

\hline
\end{tabular}
\end{table}

\section{Experiments and Results}
\subsection{Data Preprocessing and Implementation Details}
\label{sec:data}
In synthetic grids, extreme draws such as values near the limits of the range, can yield implausible voltages even when NR converges. We therefore remove outliers with a Tukey interquartile-range filter \cite{tukey1977eda} applied to bus-voltage magnitudes in per unit from the NR reference; samples outside $[\,Q_{1}-1.5\,\mathrm{IQR},\,Q_{3}+1.5\,\mathrm{IQR}\,]$ are discarded. After filtering, all quantities are converted to per unit using regime-dependent bases $V_{\text{base}}$ and $S_{\text{base}}$ with $Y_{\text{base}}=S_{\text{base}}/V_{\text{base}}^{2}$. The corpus contains $33{,}182$ HV and $31{,}500$ MV scenarios for $N=4\text{--}32$, split $1\!:\!1\!:\!1$ into train, validation, and test.

PIGNN-MLP uses 4 aggregation channels and an update MLP with hidden size 16. PIGNN-Attn uses the same hidden size with 4 attention heads and 1 attention layer per correction step; the remaining settings are identical. Line search uses initial step $\alpha_{0}=1.0$, Armijo constant $c_{1}=10^{-4}$, shrink factor $\rho=0.5$, and minimum step $\alpha_{\min}=5\times10^{-2}$. Per-iteration caps are $\Delta\theta_{\max}=0.3$~rad and $\Delta|V|_{\max}=10\%$ of the current magnitude, with voltage bounds $V_{\min}=0.8$ and $V_{\max}=1.2$. Training is unsupervised with discounted physics loss $\gamma=0.9$, $K=40$ message-passing steps, AdamW with weight decay $10^{-3}$, cosine learning-rate annealing from $10^{-4}$ to $10^{-6}$ every 20 epochs, and batch size 64 with block-diagonal batching; runs use a single RTX~3070 8\,GB GPU and an Intel Core i7–12700 with 12 cores and 20 threads, up to 4.9\,GHz. Hyperparameters were selected by a compact grid search that varied aggregation channels at 4, 8, 16, 32; hidden size from 16 to 128; message-passing steps from 10 to 80 in steps of 10; attention heads 4 or 8; attention layers 1 to 4; angle caps 0.3 to 3.14~rad; magnitude caps 10\% to 100\% of $V$; and two line search variants: steeper decrease with $\alpha_{0}=1.0$, $c_{1}=10^{-3}$, $\rho=0.5$, $\alpha_{\min}=5\times10^{-2}$; conservative monotone with $\alpha_{0}=0.5$, $c_{1}=10^{-4}$, $\rho=0.7$, $\alpha_{\min}=5\times10^{-2}$. The reported configuration is best by validation physics loss; wall-clock time breaks ties.

\subsection{Ablation Studies}
\label{sec:ablation}
\begin{enumerate}[leftmargin=1.3em]
\item \textbf{Aggregator (MLP vs.\ self-attention).}
Across HV, MV, and combined HV+MV cases, the edge-aware PIGNN-Attn consistently achieves lower voltage-magnitude errors than PIGNN-MLP. For angle predictions, PIGNN-MLP may perform slightly better in the unstabilized setting; however, once stabilization mechanisms (step caps and/or line search) are applied, PIGNN-Attn emerges as the overall best performer.

\item \textbf{Per-iteration voltage update caps.}
Step caps consistently improve voltage-magnitude accuracy but can degrade angle performance in the MLP setting. In contrast, with the Attn, caps are either beneficial or neutral and effectively suppress worst-case deviations. When combined with line search, caps no longer reduce angle accuracy and instead enhance overall robustness.

\item \textbf{Backtracking line search (LS).}
LS is the primary driver of both accuracy and robustness for both aggregators. On MV cases, Attn+LS achieves the best performance, while on HV and HV+MV cases, Attn with caps+LS is strongest overall. For the MLP, however, adding caps on top of LS provides little benefit and can even slightly degrade angle accuracy.
\end{enumerate}

For HV+MV, using smaller batches improves accuracy, by reducing cross-regime interference from mixed block-diagonal batching. Although trained on \(N\!=\!4\text{--}32\) grids, the model scales effectively to much larger dense HV grids (e.g., up to \(N=1024\)) with comparable RMSE (about \(V\!\approx\!5{\times}10^{-4}\), \(\theta\!\approx\!10^\circ\)). 

% Future work will validate scalability on larger MV networks with high \(R/X\) ratios, where strong coupling can lead to divergence, as well as on public benchmark datasets and real operational grids.
% The test result on bigger size grid and more benchmark dataset will be done in the future work as well as the more extensive existing g model comparison with GCN, Typed GNN and GAT style. and more MVN cases which can easily diverged on NR because of high coupling of Power and voltage because of high R/X ratio. and apply it to real operation grid will be tested

\begin{table}[ht]
\centering
\caption{Test RMSE of Voltage Magnitude ($V$, p.u.) and Angle ($\theta$, $^\circ$) across HV, MV, and HV+MV }
\label{tab:comparison}
\begin{tabular}{lcccccc}
\toprule
 & \multicolumn{2}{c}{\textbf{HV}} & \multicolumn{2}{c}{\textbf{MV}} & \multicolumn{2}{c}{\textbf{HV+MV}} \\
\cmidrule(lr){2-3} \cmidrule(lr){4-5} \cmidrule(lr){6-7}
\textbf{Model} & \textbf{$V$} & \textbf{$\theta$} & \textbf{$V$} & \textbf{$\theta$} & \textbf{$V$} & \textbf{$\theta$} \\
\midrule
\multicolumn{7}{c}{\textit{Base Models}} \\
\midrule
PIGNN-MLP        & 6.6e-2 & 0.62 & 7.9e-2 & 0.67 & 6.9e-2 & 1.31 \\
PIGNN-Attn       & 8.0e-3 & 0.67 & 1.5e-2 & 0.63 & 2.0e-3 & 1.27 \\
\midrule
\multicolumn{7}{c}{\textit{+Voltage Update Caps}} \\
\midrule
PIGNN-MLP        & 1.0e-3 & 8.20 & 3.9e-2 & 3.71 & 2.6e-2 & 2.88 \\
PIGNN-Attn       & 5.4e-4 & 0.16  & 1.2e-2 & 0.58 & 6.0e-3 & 1.48 \\
\midrule
\multicolumn{7}{c}{\textit{+ Line Search}} \\
\midrule
PIGNN-MLP        & 5.8e-4 & 0.26 & 1.6e-3 & 1.38 & 1.9e-3 & 1.53 \\
PIGNN-Attn       & 4.6e-4 & 0.12 & \textbf{5.7e-4} & \textbf{0.34} & 2.4e-3 & 1.26 \\
\midrule
\multicolumn{7}{c}{\textit{+Voltage Update Caps, Line Search}} \\
\midrule
PIGNN-MLP        & 5.4e-4 & 0.30 & 1.7e-3 & 1.45 & 2.1e-3 & 1.68 \\
PIGNN-Attn       & \textbf{3.3e-4} & \textbf{0.08} & 6.4e-4 & 0.41 & \textbf{1.9e-3} & \textbf{1.14} \\
\bottomrule
\end{tabular}
\end{table}

\subsection{Throughput}
We benchmark our PIGNN, implemented in PyTorch, against a Newton--Raphson baseline coded as an optimized, vectorized NumPy solver with an analytically assembled Jacobian, across HV grid sizes \(N=4\) to \(1024\) under two regimes. In the single-scenario regime, one case per \(N\) is solved: PIGNN runs on a GPU with batch size 1, and NR runs on a CPU with one worker. In the multi-scenario regime, 4096 cases per \(N\) are solved: NR uses 20 CPU workers, and PIGNN runs on a single 8\,GB GPU with streaming micro-batches that fully utilize device memory without overflow. To avoid thread oversubscription in CPU multiprocessing, BLAS threads are set to one per process. Timings use the hardware and software setup described in Section~\ref{sec:data}. Across both regimes, NR time scales super-linearly but polynomially with \(N\) due to sparse factorization and fill-in, while PIGNN grows more gently. In single-scenario runs, NR is competitive for \(N<200\), but PIGNN-Attn-LS becomes faster as \(N\) increases; between \(N=256\) and \(N=896\) it is about \(2\times\) faster on average, with peak speedups of \(4\text{--}5\times\) at the largest grids. In multi-scenario runs, PIGNN-Attn-LS achieves about \(3\times\) higher throughput than NR on average and is consistently faster across all grid sizes.

 % Our results show that while NR remains competitive for single, small cases, PIGNN delivers substantial throughput gains in the multi-scenario setting by amortizing GPU kernels across scenarios, making it attractive for contingency screening and stochastic studies.

% Multi-scenario PF reflects real operations where many independent PFs must be solved---e.g., N-1 RTCA runs a PF per credible outage \cite{Tang2018RTCA}, and probabilistic planning/forecasting evaluates large Monte-Carlo ensembles \cite{Carpinelli2015MLMC,Kabir2016PVPLF}. 

\begin{figure}[t]
  \centering
  \includegraphics[width=\linewidth]{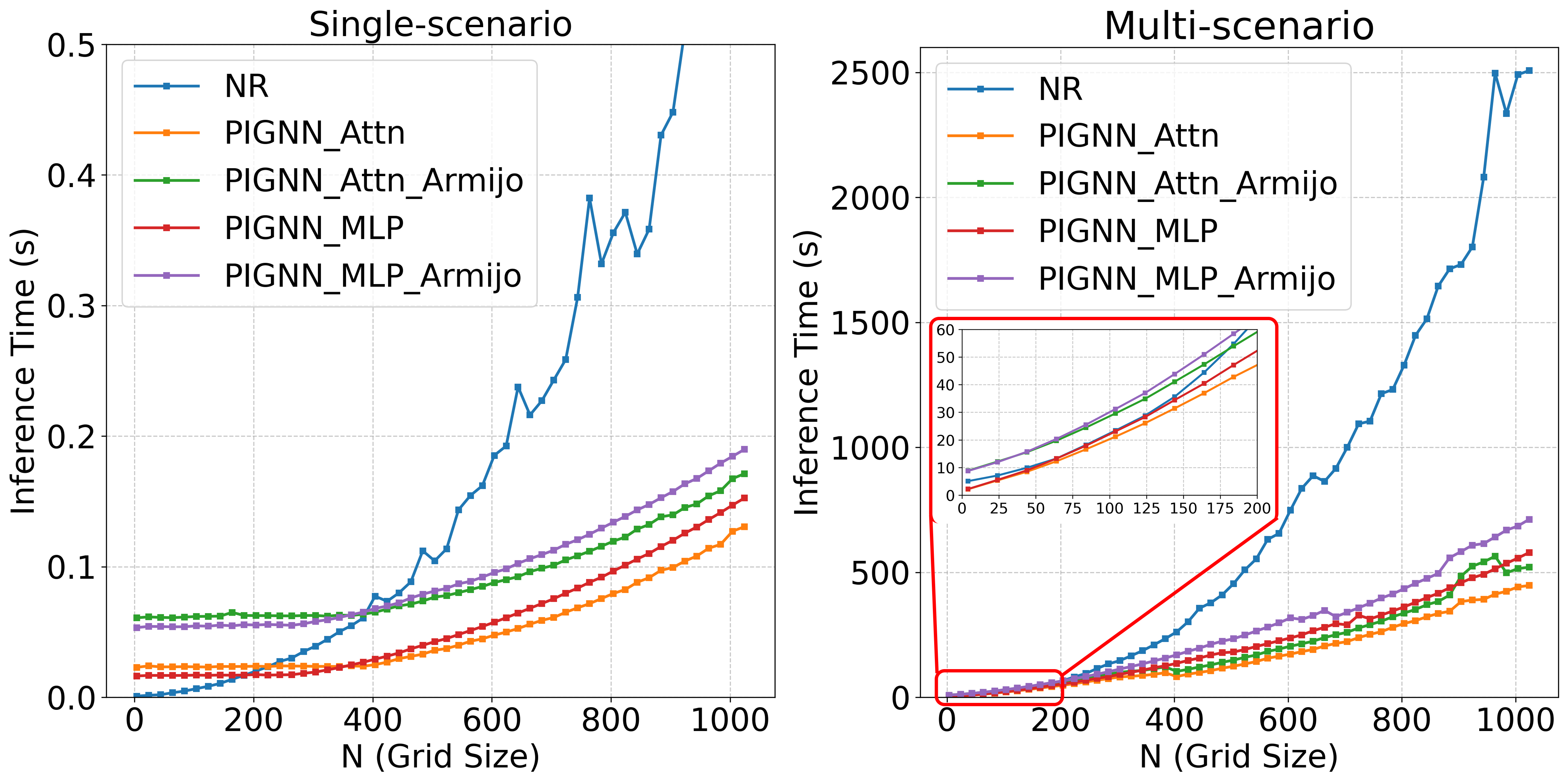}
    \caption{Computational time over grid size for NR and PIGNN variants. Results are reported as median inference time (over multiple runs with 2 warmup and 5 repeat cycles).}
  \label{fig:time_vs_grid}
\end{figure}

% d_hi=16
% GNSMsg_armijo         params=90,640
% GNSMsg_EdgeSelfAttn   params=7,508

% d_hi=32
% GNSMsg_armijo         params=262,800
% GNSMsg_EdgeSelfAttn   params=20,852

% d_hi=32, K= 10
% GNSMsg_armijo                params=65,700
% GNSMsg_EdgeSelfAttn          params=14,912

\section{Conclusion}

% This work proposes a physics-informed GNN with edge-aware attention and a line-search--based globalized correction operator for HV/MV AC power flow, trained without Newton--Raphson supervision. The model closely matches NR and outperforms a PIGNN-MLP baseline, while delivering \(2\times\) faster single-case inference and \(\sim\!3\times\) speedups in multi-scenario batching than NR, making it a strong NR surrogate.

This work advances AC power-flow solvers by introducing a physics-informed GNN with node/edge physics injection, edge-aware attention, and a line-search--based globalized correction operator, trained without Newton--Raphson supervision. The model achieves NR-level accuracy, and delivers substantial inference speedups. These results highlight the promise of neural-network-based solvers for power systems and motivate future work on broader scalability, rigorous benchmarking, and integration into operational systems.

\bibliographystyle{IEEEbib}
\bibliography{refs}

\end{document}